\documentclass{article} 
\usepackage{times}

\usepackage{amsmath,amsfonts,bm}

\def\eqref#1{equation~\ref{#1}}

\def\1{\bm{1}}

\DeclareMathAlphabet{\mathsfit}{\encodingdefault}{\sfdefault}{m}{sl}
\SetMathAlphabet{\mathsfit}{bold}{\encodingdefault}{\sfdefault}{bx}{n}

\usepackage{hyperref}
\usepackage{url}

\usepackage{iclr2026_conference,amsmath,graphicx,hyperref}
\usepackage{array}
\usepackage[dvipsnames]{xcolor}
\usepackage{multirow}
\usepackage{booktabs}
\usepackage{etoolbox}
\usepackage{wrapfig}
\usepackage{caption}

\title{Index-Preserving Lightweight Token Pruning for Efficient Document Understanding in Vision-Language Models
}

\author{Jaemin Son \quad Sujin Choi \quad Inyong Yun\\
Hana Institute of Technology, Hana TI\\
\texttt{\{woalsdnd, sujin\_choi, iyyun\}@hanafn.com}
}

\iclrfinalcopy

\begin{document}
%
\maketitle
\begin{abstract}
Recent progress in vision-language models (VLMs) has led to strong accuracy on document understanding tasks such as parsing and key information extraction, but processing high-resolution document images remains computationally expensive. We propose a lightweight pre-encoder token pruning framework that removes non-informative background patches using a binary text-region classifier with a max-pooling refinement step. The framework preserves token indices to maintain the spatial correspondence required for layout-sensitive recognition. Experiments on real-world document benchmarks show 40–60\% FLOPs reduction while maintaining comparable accuracy.

\end{abstract}

\section{Introduction}
\label{sec:intro}

Recent advancements in vision-language models (VLMs)~\citep{team2024gemini, bai2025qwen2,grattafiori2024llama,touvron2023llama,achiam2023gpt,lu2024deepseek,beyer2024paligemma} have broadened their applications in document understanding tasks, including document layout parsing~\citep{yang2024cc}, key information extraction~\citep{park2019cord}, and visual question answering~\citep{mathew2021docvqa}. By jointly modeling text and vision features, these models achieve strong performance with minimal task-specific tuning. However, their high computational cost remains a critical obstacle to practical deployment.

To address this challenge, token pruning and merging techniques have been proposed in the vision domain~\citep{kim2024token, rao2021dynamicvit, bolya2022token, zheng2023less, kong2022spvit}. These methods reduce redundant visual tokens, improving inference efficiency with minimal accuracy degradation. Yet, such strategies are underexplored in document understanding tasks, where large background regions allow aggressive filtering without affecting recognition performance.

In this work, we introduce a lightweight and effective token pruning method tailored for document understanding, aiming to reduce unnecessary computation while preserving accuracy\footnote{\url{https://github.com/jaeminSon/
index-preserving-lightweight-token-pruning}}. In summary, the key findings are as follows:

\hangindent=1.8em
\textbf{• Lightweight Patch Classification} effectively filters out background regions in document images, enabling substantial computational savings.

\hangindent=1.8em
\textbf{• Preserving the original indices} of the remaining tokens after pruning has a significant impact on performance in document understanding tasks.

\section{Related work}
\label{sec:related_work}

Existing methods for ViTs aim to improve efficiency by removing or merging uninformative tokens in classification and detection tasks. DynamicViT~\citep{rao2021dynamicvit} inserts lightweight prediction modules between transformer blocks to dynamically determine which tokens to retain. Focus-DETR~\citep{zheng2023less} selects foreground-relevant tokens from multi-scale feature maps for object detection, pruning away background regions early. SPViT~\citep{kong2022spvit} compresses uninformative tokens into a single package token to retain global context while reducing count. ToMe~\citep{bolya2022token} merges spatially and semantically similar tokens using cosine similarity in each transformer block, while Token Fusion~\citep{kim2024token} modifies this by weighted merging that preserves magnitude to retain more informative content.

In document understanding, DocKylin~\citep{zhang2025DocKylin} applies Sobel filtering to remove white backgrounds and merges tokens in a language embedding space. Unlike the prior work, our method prunes before the vision encoder to maximize the effectiveness of pruning. Additionally, we retain the spatial indices of the remaining tokens, which is crucial for processing pruned tokens within the vision encoder.

\begin{figure*}[t]
\centering
\centerline{\includegraphics[width=\textwidth]{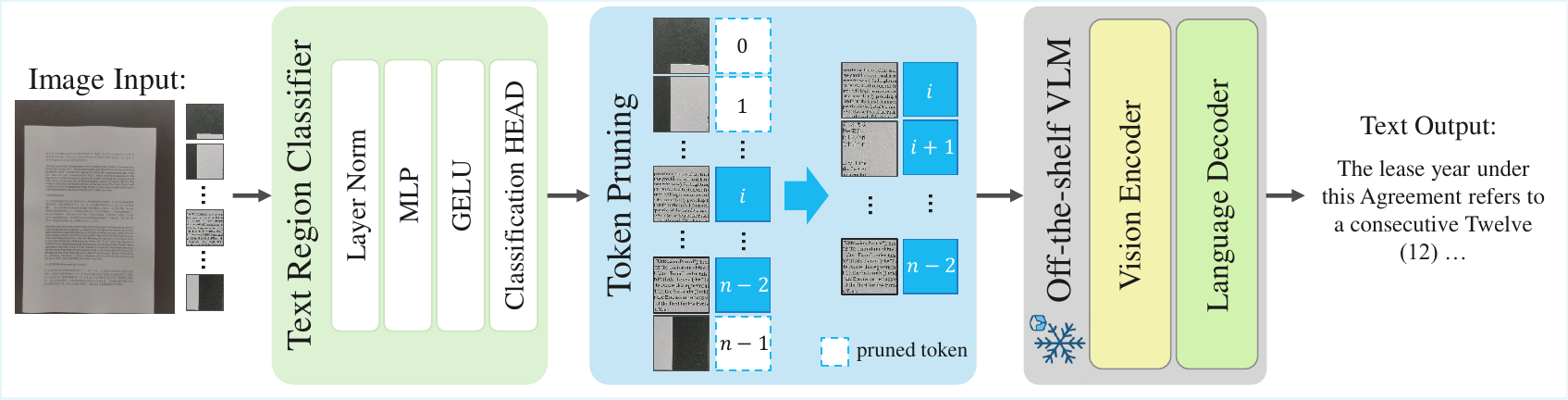}}
\caption{\textbf{Illustration of the Index-Preserving Lightweight Token Pruning framework.} The proposed framework consists of three components: a binary text-region classifier (green), index-preserving token pruning (blue) and a frozen off-the-shelf VLM (gray). Text-region patches are selected and fed into an off-the-shelf VLM.}
\label{fig:main}
\end{figure*}

\section{Method}
\label{sec:method}
We propose an index-preserving pruning method that discards background patches at the earliest stage, before any computation occurs in a VLM. See Figure~\ref{fig:main} for the overview.

\subsection{Lightweight text-region classifier}
\label{ssec:classifier}

A lightweight binary classifier operates on image patches to predict whether each contains text (foreground) or can be discarded (background). Unlike existing methods, our approach applies pruning prior to any visual encoding or language decoding. This eliminates non-informative regions early, reducing the input size for computationally expensive components.

\subsection{Token pruning with index preservation}
To maintain spatial coherence, we preserve the original indices of the selected patches after pruning. Each patch is paired with its index, and those recognized as foreground text are forwarded to the VLM along with their original indices. Maintaining the original indices is crucial for document understanding tasks, where positional information encodes critical semantics such as text content and spatial layout. Without index preservation, the decoder receives irregularly stitched patches with misaligned positional information, which severely degrades text recognition.

\subsection{Foreground refinement with max-pooling}

\begin{wrapfigure}{r}{0.45\textwidth}
  \centering
  \vspace{-15pt} 
    \includegraphics[height=3cm]{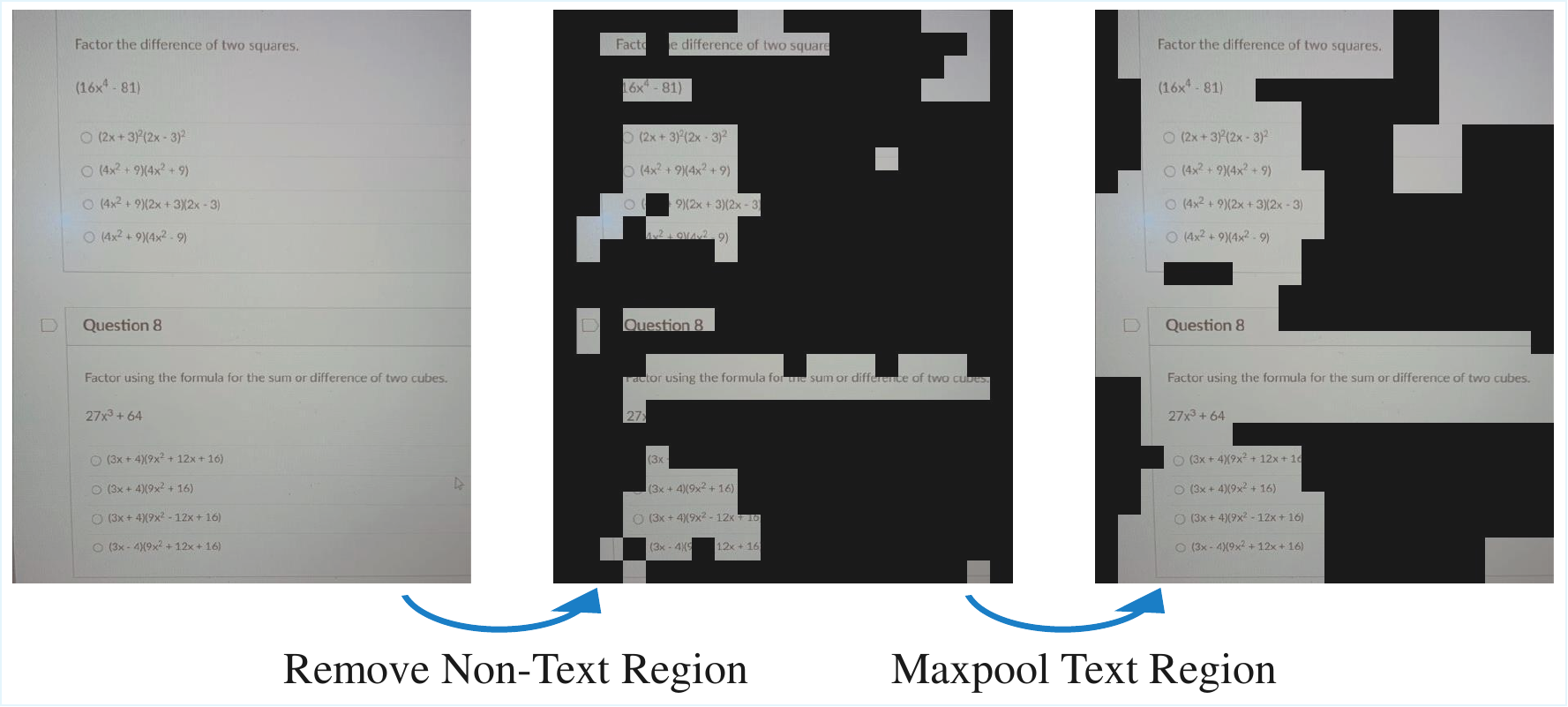}
    \caption{\bf{Text recovery with max-pooling.}}
    \label{fig:processing}
  \vspace{-20pt}
\end{wrapfigure}

Patch-level classification often produces fragmented foreground masks, where parts of text regions are missed. This is because the classifier does not explicitly account for spatial continuity.

To address this, we apply a max-pooling operation to the binary mask, which enhances spatial continuity and recovers adjacent missed regions. This refinement significantly improves the coherence of selected text areas, as illustrated in Figure~\ref{fig:processing}.

\section{Experiments}
\label{sec:majhead}

\subsection{Evaluation setup}
\label{subsec:vlm_setup}

We evaluated our pruning approach (Section~\ref{ssec:implementation}) with Qwen2.5-VL on CC-OCR dataset~\citep{yang2024cc}. We retained the evaluation logic in the official implementations~\citep{cc-ocr-code} and focused on English documents with ample marginal background — particularly, document parsing tasks (LaTeX formulas and HTML tables) and key information extraction tasks (receipts).

Two images in the CORD subset were fully pruned, meaning all patches were removed by the classifier. We excluded them from the FLOPs computation and assigned 0 score for F1 and the normalized TED accuracy.

\begin{figure*}[t]
\centering

\begin{minipage}{0.3\textwidth}
    \centering
    \includegraphics[width=\linewidth]{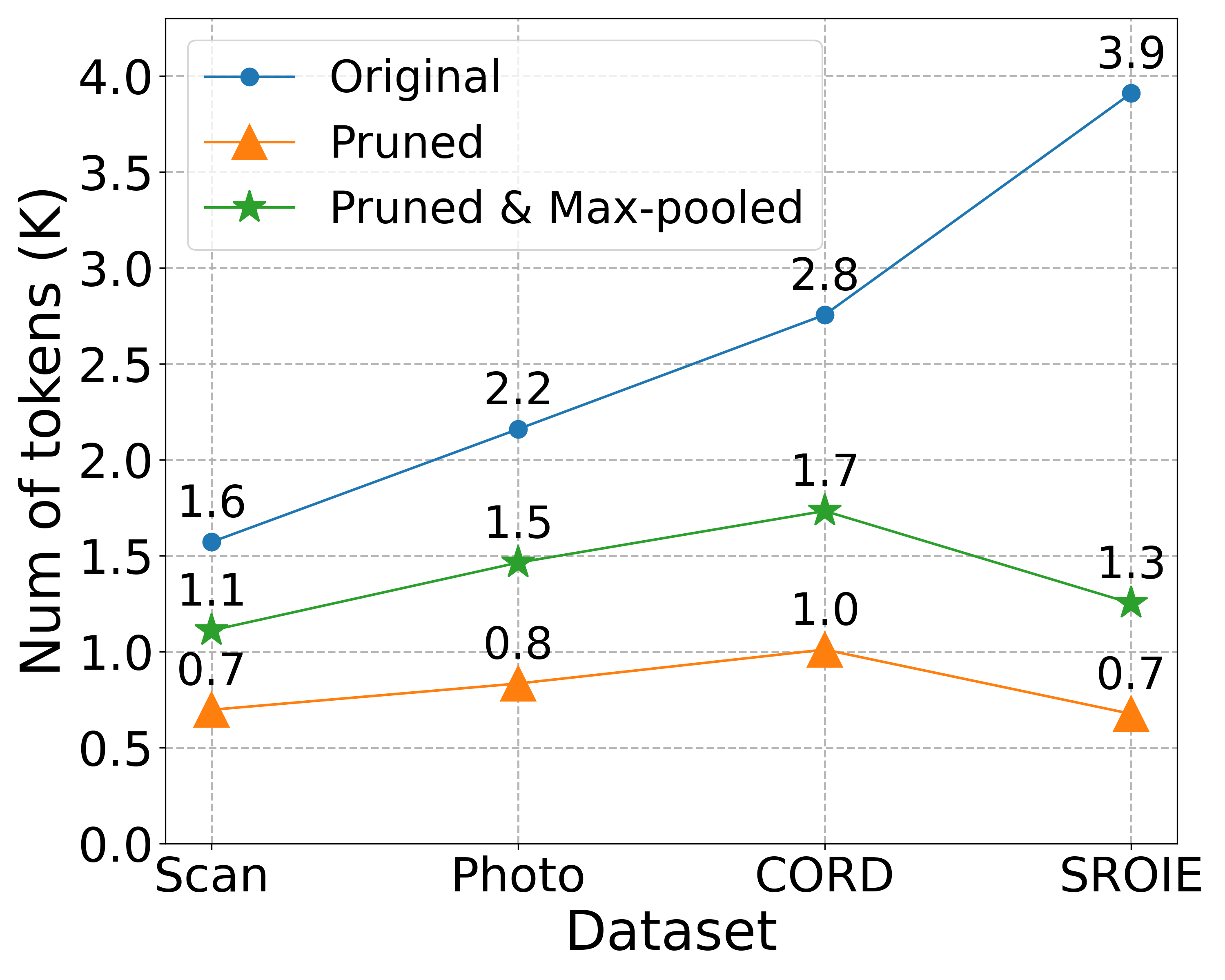}
    \centerline{(a) Visual token count.}\medskip

\end{minipage}
\hfill
\begin{minipage}{0.3\textwidth}
    \centering
    \includegraphics[width=\linewidth]{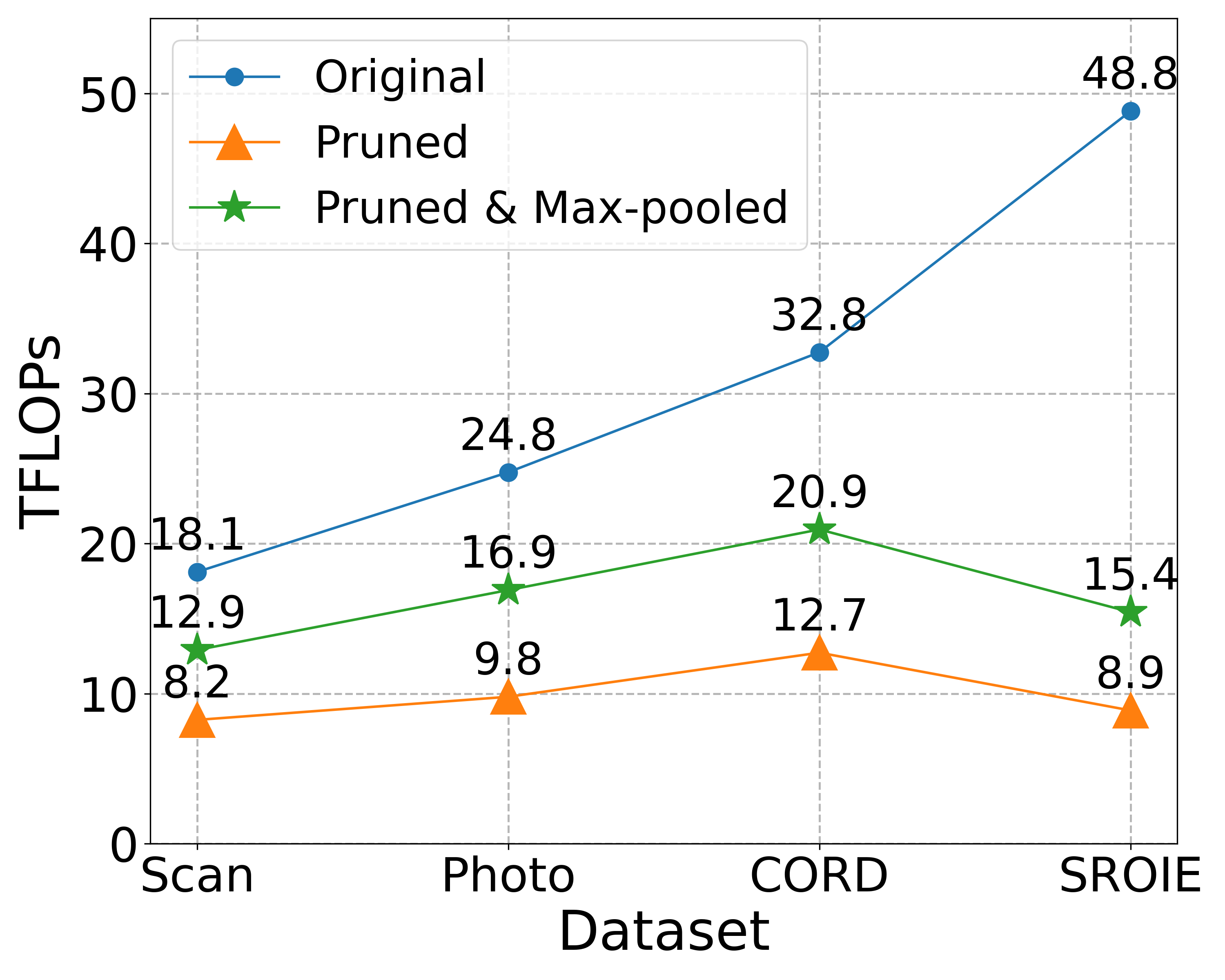}
    \centerline{(b) TFLOPs for Qwen2.5-VL 3B.}\medskip
\end{minipage}
\hfill
\begin{minipage}{0.3\textwidth}
    \centering
    \includegraphics[width=\linewidth]{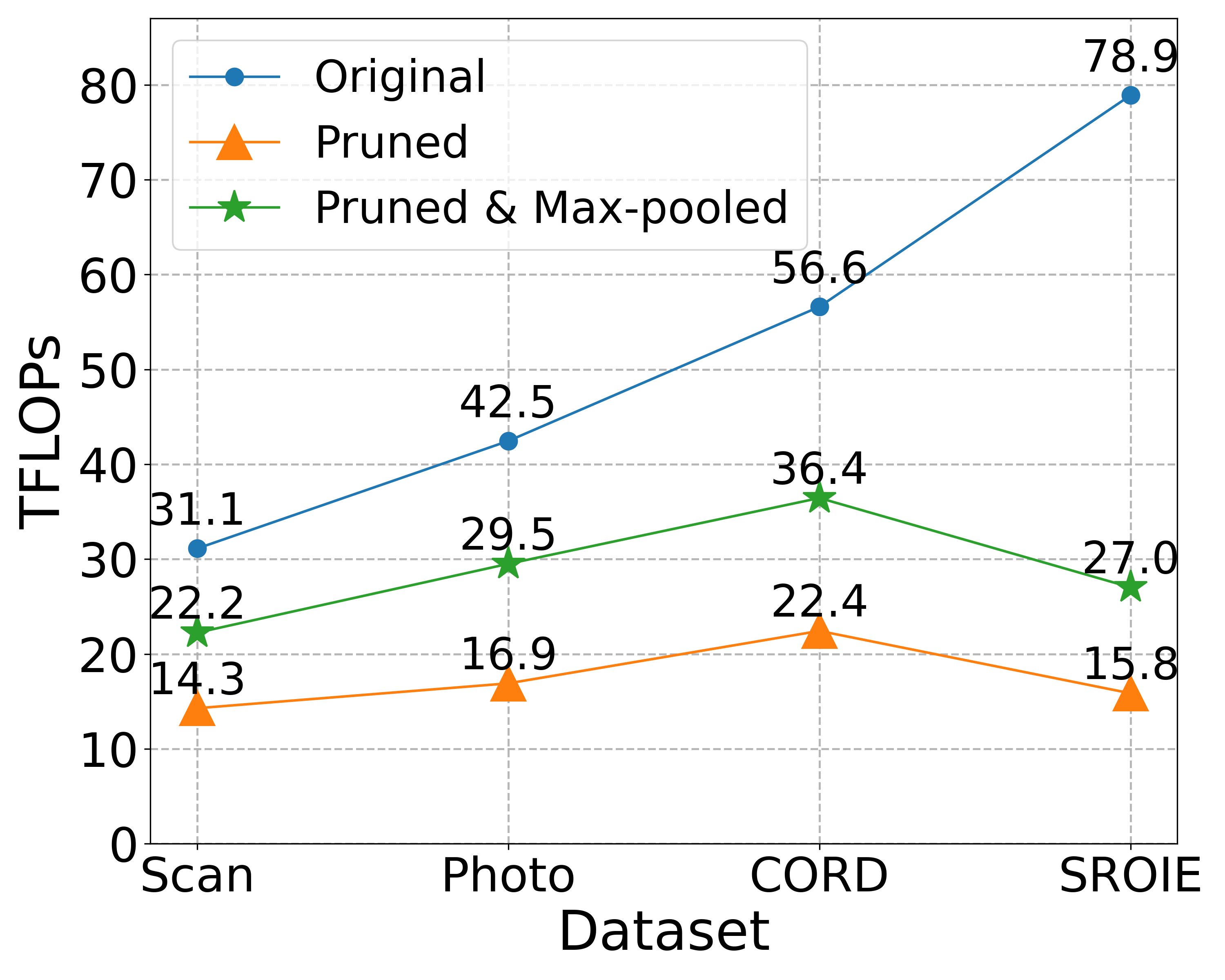}
    \centerline{(c) TFLOPs for Qwen2.5-VL 7B.}\medskip

\end{minipage}
\caption{ \textbf{Visual token and TFLOPs statistics}. Metrics are averaged over all images in each dataset. TFLOPs are counted end-to-end including the computation of the text-region classifier.}

\label{fig:count_token_FLOPs}
\end{figure*}

\subsection{Token and FLOPs reduction}
\label{subsec:flops}

As shown in Figure~\ref{fig:count_token_FLOPs}(a), token pruning alone reduced the number of visual tokens by an average of 65.7\% across datasets, while the combination of pruning and max-pooling yielded a reduction of 41.6\%. Max-pooling expands the foreground mask, retaining additional neighboring patches. Receipt-style images often have abundant backgrounds and smaller text regions compared to document images. This allows for more aggressive token pruning in CORD and SROIE datasets.

We measured end-to-end Tera Floating-point Operations (TFLOPs) using the Calflops Python module~\citep{ye2023calflops}. FLOPs were computed over the forward pass of the complete VLM pipeline and averaged across all images in each dataset. Due to memory constraints, we present results for the 3B and 7B variants only, as shown in Figure~\ref{fig:count_token_FLOPs}(b) and ~\ref{fig:count_token_FLOPs}(c), respectively.

Our pruning approach consistently achieved over 60\% FLOPs reduction. On the SROIE dataset, FLOPs dropped by $\sim$80\%. When combined with max-pooling, FLOPs are reduced by 40–60\% on all datasets.

\begin{table*}[t]
\renewcommand{\arraystretch}{0.8}
\centering

\resizebox{12cm}{!}{%
\begin{tabular}{
    >{\centering\arraybackslash}p{0.08\textwidth}
    !{\vrule width 1pt}
    >{\raggedright\arraybackslash}p{0.3\textwidth}
    !{\vrule width 1pt}
    >{\raggedleft\arraybackslash}p{0.07\textwidth}
    >{\raggedleft\arraybackslash}p{0.07\textwidth}@{\hskip 15pt}
    !{\vrule}
    >{\raggedleft\arraybackslash}p{0.06\textwidth}
    >{\raggedleft\arraybackslash}p{0.06\textwidth}@{\hskip 20pt}
    >{\raggedleft\arraybackslash}p{0.07\textwidth}
    >{\raggedleft\arraybackslash}p{0.06\textwidth}@{\hskip 15pt}
}
\toprule
\multirow{3}{*}{Params} 
  & \multirow{3}{*}{\parbox[c][2.5em][c]{\linewidth}{\centering Method}} 
  & \multicolumn{2}{c!{\vrule width 0.7pt}}{Document Parsing} 
  & \multicolumn{4}{c}{Key Information Extraction} \\
\cmidrule(lr){3-4} \cmidrule(lr){5-8}
  & & Scan & Photo  & \multicolumn{2}{c}{CORD} & \multicolumn{2}{c}{SROIE} \\
\cmidrule(lr){3-3} \cmidrule(lr){4-4} \cmidrule(lr){5-6} \cmidrule(lr){7-8}
  & & ANLS & ANLS & F1 & Acc & F1 & Acc \\
\midrule
\multirow{3}{*}{3B} 
  & Original & 62.4 & 73.7  & 87.2 & 94.7  & 88.7 & 97.5  \\
  & \hspace{10pt} Pruned ($\Delta$)  
    & \textcolor{gray}{-13.6} & \textcolor{gray}{-13.0}
    & \textcolor{gray}{-30.8} & \textcolor{gray}{-17.1}
    & \textcolor{gray}{-15.4} & \textcolor{gray}{-2.7} \\
  & \hspace{10pt} Pruned \& Max-pooled ($\Delta$) 
    & \textcolor{Green}{-0.6} & \textcolor{Green}{-2.7}
    & \textcolor{Green}{-4.2} & \textcolor{Green}{-4.4}
    & \textcolor{Green}{-0.8} & \textcolor{blue}{+0.3} \\
\midrule
\multirow{3}{*}{7B} 
  & Original & 64.7 & 69.9 & 89.5 & 96.3 & 90.7 & 98.1  \\
  & \hspace{10pt} Pruned ($\Delta$)  
    & \textcolor{gray}{-11.0} & \textcolor{gray}{-10.2}
    & \textcolor{gray}{-24.7} & \textcolor{gray}{-15.9}
    & \textcolor{gray}{-13.7} & \textcolor{gray}{-1.9} \\
  & \hspace{10pt} Pruned \& Max-pooled ($\Delta$) 
    & \textcolor{Green}{-2.7} & \textcolor{Green}{-1.3}
    & \textcolor{Green}{-5.3} & \textcolor{Green}{-4.8}
    & \textcolor{Green}{-0.4} & \textcolor{blue}{+0.1} \\
\midrule
\multirow{3}{*}{32B} 
  & Original & 60.9  & 67.5  & 87.0 & 95.2  & 88.5 & 97.5  \\
  & \hspace{10pt} Pruned ($\Delta$)  
    & \textcolor{gray}{-9.5} & \textcolor{gray}{-16.6}
    & \textcolor{gray}{-26.0} & \textcolor{gray}{-14.8}
    & \textcolor{gray}{-14.4} & \textcolor{gray}{-2.3} \\
  & \hspace{10pt} Pruned \& Max-pooled ($\Delta$) 
    & \textcolor{blue}{+1.7} & \textcolor{Green}{-1.9}
    & \textcolor{Green}{-3.3} & \textcolor{Green}{-4.0}
    & \textcolor{Green}{-0.0} & \textcolor{blue}{+0.1} \\
\midrule
\multirow{3}{*}{72B} 
  & Original & 67.7 & 70.0 & 92.8 & 97.6 & 91.6 & 98.6 \\
  & \hspace{10pt} Pruned ($\Delta$)  
    & \textcolor{gray}{-7.0} & \textcolor{gray}{-4.5}
    & \textcolor{gray}{-24.3} & \textcolor{gray}{-13.5}
    & \textcolor{gray}{-13.8} & \textcolor{gray}{-1.8} \\
  & \hspace{10pt} Pruned \& Max-pooled ($\Delta$) 
    & \textcolor{Green}{-0.4} & \textcolor{blue}{+1.5}
    & \textcolor{Green}{-3.8} & \textcolor{Green}{-4.0}
    & \textcolor{Green}{-0.5} & \textcolor{Green}{-0.1} \\

\bottomrule
\end{tabular}%
}
\caption{ \textbf{Performance degradation due to pruning and the effect of max-pooling.} Evaluation results are shown for the original Qwen2.5-VL models on document understanding tasks in the CC-OCR dataset. Relative differences ($\Delta$) are shown for the pruned variants with respect to the original model of the same parameter size. Acc represents normalized TED accuracy.}
\label{tab:result}
\end{table*}

\subsection{Impact of pruning and max-pooling}
\label{ssec:subhead}

As shown in Table~\ref{tab:result}, applying text-region pruning alone led to a noticeable degradation in model performance. In document parsing tasks, performance dropped up to $\sim$17\%p in ANLS. Similarly, key information extraction datasets showed severe decreases in F1-score, up to $\sim$31\%p, underscoring the sensitivity of downstream models to missing text regions.

Introducing max-pooling after pruning mitigated much of this degradation. In document parsing tasks, ANLS scores remained largely unaffected after max-pooling, showing at most $\sim$3\%p decrease. In some cases, performance even improved, possibly because the model focused more on informative regions. This indicates that max-pooling can effectively recover some of the missed regions discarded by the classifier. 

Similar patterns emerged in key information extraction tasks. On SROIE, F1-score and the accuracy remained nearly unchanged after applying pruning and max-pooling. In CORD, minor decreases of up to $\sim$5\%p were observed, which is a significant improvement over the pruning-only results.

\subsection{Comparison with existing methods}
\label{subsec:comp}

\begin{wraptable}{r}{0.5\textwidth}
\scalebox{0.8}{
\begin{tabular}{
    >{\raggedright\arraybackslash}p{0.1\textwidth}
    !{\vrule}
    >{\raggedleft\arraybackslash}p{0.05\textwidth}
    >{\raggedleft\arraybackslash}p{0.01\textwidth}
    !{\vrule}
    >{\centering\arraybackslash}p{0.13\textwidth}
    >{\centering\arraybackslash}p{0.13\textwidth}
}
\toprule
\multirow{2}{*}{\parbox[c][1.5em][c]{\linewidth}{\centering Method}} 
  & \multicolumn{2}{c!{\vrule}}{Doc. Parsing} 
  & \multicolumn{2}{c!}{Key Info. Extract.}  \\
\cmidrule(lr){2-3} \cmidrule(lr){4-5}
  & {\centering Scan} & {\centering Photo} & {\centering CORD} & {\centering SROIE} \\
\midrule
Original & 62.4 & 73.7  & 87.2 (94.7)  & 88.7 (97.5) \\
\midrule
ToME & 8.8&11.1& 6.0 (13.5)& 0.0 (9.9) \\
DocKylin & 34.3 &47.7& 73.1 (84.1)& 69.6 (83.5)  \\
\textbf{Ours} & \textbf{61.8} & \textbf{71.0} & \textbf{83.0 (90.3)} & \textbf{87.9 (97.8)}  \\

\bottomrule
\end{tabular}%
}
\caption{
{\bf Comparison with existing methods.} Metrics are ANLS for document parsing and F1 and normalized TED accuracy for key information extraction.}
\label{tab:comp}
\vspace{-10pt}
\end{wraptable}
We integrated official implementations of existing methods in Qwen2.5-VL-Instruct-3B. For ToMe~\citep{bolya2022token}, we merged 1\% of tokens at each ViT layer. For DocKylin~\citep{zhang2025DocKylin}, we integrated only Dynamic Token Slimming (DTS) excluding Adaptive Pixel Slimming, which produces a new compact image.

As shown in Table~\ref{tab:comp}, ToMe~\citep{bolya2022token} showed low ANLS and accuracy. As it shuffles and rearranges tokens at each merge step, the index structure collapses leading to poor text recognition. This suggests that, unlike classification tasks, text recognition requires proper token indexing to attain reasonable performance. Also, DocKylin~\citep{zhang2025DocKylin} with DTS showed limited performance. Its merging strategy treats tokens highly correlated with others as background, under the assumption that backgrounds share similar visual patterns. When this assumption does not hold, the method becomes less effective.

\subsection{Ablation on indexing strategy}

\begin{wraptable}{r}{0.5\textwidth}
\vspace{-5mm}
\scalebox{0.8}{
  \begin{tabular}{
    >{\raggedright\arraybackslash}p{0.1\textwidth}
    !{\vrule}
    >{\raggedleft\arraybackslash}p{0.05\textwidth}
    >{\raggedleft\arraybackslash}p{0.02\textwidth}
    !{\vrule}
    >{\centering\arraybackslash}p{0.12\textwidth}
    >{\centering\arraybackslash}p{0.12\textwidth}
}
\toprule
\multirow{2}{*}{\parbox[c][1.5em][c]{\linewidth}{\centering Indexing Strategies}} 
  & \multicolumn{2}{c!{\vrule}}{Doc. Parsing} 
  & \multicolumn{2}{c}{Key Info. Extract.}\\
\cmidrule(r){2-3} \cmidrule(r){4-5}
  & {\centering Scan} & {\centering Photo} & {\centering CORD} & {\centering SROIE} \\
\midrule

Constant  &  9.1 &  5.8 & 3.5 (19.7) &  0.0 (10.9)  \\
Random    & 16.0 & 13.7 & 8.5 (27.9) & 0.2 (11.7)  \\
Ordered   & 36.2 & 49.2 & 38.8 (58.0)& 40.7 (65.1)  \\
\textbf{Preserved} & \textbf{61.8} & \textbf{71.0} & \textbf{83.0 (90.3)} & \textbf{87.9 (97.8)}  \\

\bottomrule
\end{tabular}%
}
\caption{ \textbf{Ablation of indexing strategies.} Metrics are ANLS for document parsing and F1 and normalized TED accuracy for key information extraction.}
\label{tab:indexing}
\vspace{-10mm}
\end{wraptable}

We tested alternative indexing strategies using Qwen2.5-VL-Instruct-3B: setting all indices to zero (constant), assigning indices randomly (random), or incrementally assigning integers from 0 to $L-1$ (ordered).

Table~\ref{tab:indexing} shows that constant and random indexing perform poorly, and ordered indexing remains far below preserving the original indices, indicating that index preservation is essential for maintaining spatial correspondence.

\section{Conclusion}
\label{sec:conclusion}

We proposed a straightforward yet effective token pruning strategy tailored for vision-language models in document understanding tasks. Across multiple document text datasets, pruning and max-pooling substantially reduce computation with only minor performance degradation. These findings demonstrate the promise of early-stage pruning for efficient and effective document analysis.

\vfill\pagebreak



\bibliography{iclr2026_conference}
\bibliographystyle{iclr2026_conference}

\vfill\pagebreak

\appendix
\section{Appendix}

\subsection{Implementation details}
\label{ssec:implementation}

We implemented our method using Qwen2.5-VL-Instruct~\citep{bai2025qwen2} of various sizes, available via Hugging Face Transformers module~\citep{wolf-etal-2020-transformers}. Qwen2.5-VL models accept variable-length image tokens without resizing input images, enabling seamless integration of our index-preserving pruning. This flexibility is typically limited in fixed-length models like PaliGemma~\citep{steiner2024paligemma}. Furthermore, unlike DINOv2~\citep{darcet2023vision}, Qwen2.5-VL does not rely on register tokens in transformer layers within the vision encoder, which are prone to performance degradation when the number of tokens is reduced. 

After training the classifier with binary cross entropy loss, we prepended it to the vision encoder. Image patches with classifier logits greater than zero are retained as text regions and max-pooled by $3\times 3$. Then, the indices of the selected tokens were preserved and subsequently passed to the vision encoder and the language decoder. We experimented with Bfloat16 precision and flash-attention-2~\citep{dao2023flashattention}. We set the maximum generation length to 2048 tokens, which we found sufficient in our experimental settings.

Note that our lightweight classifier contains a minimal number of parameters, enabling fine-tuning even with minimal resources.

\subsection{Patch classifier dataset}
\label{ssec:patch_dataset}

We sampled 800 OCR document images from AI-Hub\footnote{This research used datasets from The Open AI Dataset Project (AI-Hub, S. Korea). All data information can be accessed through AI-Hub (\url{https://www.aihub.or.kr})}. Using PSENet~\citep{li2020psenet}, we extracted text bounding boxes and assigned binary labels to square image patches: foreground (text) if the patch overlapped any bounding box, and background (non-text) otherwise. Based on ablation results (Section~\ref{subsec:ablation}), we used patch size 28 for all experiments.

\subsection{Ablation on patch size}
\label{subsec:ablation}

Square image patches were generated with side lengths of 14, 28, 56, and 112 pixels. Each patch-size-specific dataset contains approximately 99,600 training samples and 400 validation samples, with a balanced split between foreground and background. Ambiguous validation images were manually removed. We fixed the linear projection dimension to 256 and evaluated the best-performing weights for each patch size.

\begin{table}[htp]
\renewcommand{\arraystretch}{0.8}
\centering

\resizebox{4cm}{!}{%

\begin{tabular}{c|rr}
\toprule
Patch Size & Params(K) & AP \\
\midrule
14   & 51    & 0.96 \\
\textbf{28}   & \textbf{203}   & \textbf{0.99} \\
56   & 810   & 0.98 \\
112  & 3000  & 0.98 \\
\bottomrule
\end{tabular}
}
\caption{Average Precision across patch sizes.}
\label{tab:patch_ap}
\end{table}

As shown in Table~\ref{tab:patch_ap}, models with patch sizes of 28 and above performed similarly, while the smallest size of 14 showed degraded performance. This can be explained by character sizes in documents. An A4 page scanned at 300 DPI (Dots Per Inch) produces a $2481 \times 3507$ pixel image. A 12-point font amounts to $\sim$50 pixels and has effective character heights of 30–35 pixels, considering spacing. A patch size of 14 is too small to capture complete character structures and may fail to distinguish texts from noise.

\end{document}